\documentclass[sigconf,nonacm, 9pt]{acmart}

\usepackage{color,soul}
\usepackage{floatflt}
\usepackage{mathtools, nccmath}
\usepackage{microtype}

\usepackage{array}
\newcolumntype{P}[1]{>{\centering\arraybackslash}p{#1}}

\newcommand{\name}{Ev-Edge}%

\newcommand{\minus}{\scalebox{0.75}[1.0]{$-$}}

\AtBeginDocument{%
  \providecommand\BibTeX{{%
    \normalfont B\kern-0.5em{\scshape i\kern-0.25em b}\kern-0.8em\TeX}}}

\begin{document}

\title{\name: Efficient Execution of Event-based Vision Algorithms on Commodity Edge Platforms
}

\author{Shrihari Sridharan}
\affiliation{%
  \institution{School of ECE, Purdue University}
  \country{} 
}
\email{sridhar4@purdue.edu}

\author{Surya Selvam}
\affiliation{%
  \institution{School of ECE, Purdue University}
  \country{} 
}
\email{selvams@purdue.edu}

\author{Kaushik Roy}
\affiliation{%
  \institution{School of ECE, Purdue University}
  \country{} 
}
\email{kaushik@purdue.edu}

\author{Anand Raghunathan}
\affiliation{%
  \institution{School of ECE, Purdue University}
  \country{} 
}
\email{raghunathan@purdue.edu}

\begin{abstract}

Event cameras have emerged as a promising sensing modality for autonomous navigation systems, owing to their high temporal resolution, high dynamic range and negligible motion blur. To process the asynchronous temporal event streams from such sensors, recent research has shown that a mix of Artificial Neural Networks (ANNs), Spiking Neural Networks (SNNs) as well as hybrid SNN-ANN algorithms are necessary to achieve high accuracies across a range of perception tasks. However, we observe that executing such workloads on commodity edge platforms which feature heterogeneous processing elements such as CPUs, GPUs and neural accelerators results in inferior performance. This is due to the mismatch between the irregular nature of event streams and diverse characteristics of algorithms on the one hand and the underlying hardware platform on the other. We propose \name{}, a framework that contains three key optimizations to boost the performance of event-based vision systems on edge platforms: (1) An Event2Sparse Frame converter directly transforms raw event streams into sparse frames, enabling the use of sparse libraries with minimal encoding overheads (2) A Dynamic Sparse Frame Aggregator merges sparse frames at runtime by trading off the temporal granularity of events and computational demand thereby improving hardware utilization (3) A Network Mapper maps concurrently executing tasks to different processing elements while also selecting layer precision by considering both compute and communication overheads. On several state-of-art networks for a range of autonomous navigation tasks, \name{} achieves 1.28x-2.05x improvements in latency and 1.23x-2.15x in energy over an all-GPU implementation on the NVIDIA Jetson Xavier AGX platform for single-task execution scenarios. ~\name{} also achieves 1.43x-1.81x latency improvements over round-robin scheduling methods in multi-task execution scenarios.

\end{abstract}



\sloppy
\maketitle
\section{Introduction}
Event cameras, also known as Dynamic Vision Sensors (DVS)~\cite{dvs}, have demonstrated great potential in the field of robotics and autonomous systems such as self-driving cars and unmanned aerial vehicles (UAVs)~\cite{eventsurvey}. These bio-inspired low-latency sensors produce asynchronous streams of brightness changes per pixel called events, offering high temporal resolution (millions of events per second), high dynamic range (140 dB) and negligible motion blur~\cite{eventsurvey}. In order to effectively process these event data streams, there is a need for suitable algorithms that operate on this sensing modality. Consequently, recent efforts ~\cite{fusionflow,halsie,dotie,depth} have shown that artificial neural networks (ANNs), spiking neural networks (SNNs) as well as hybrid SNN-ANN algorithms have excelled on a variety of event-based vision tasks such as optical flow estimation, semantic segmentation and depth estimation. 


\begin{floatingfigure}[r]{0.75\columnwidth}
  \centering
   \includegraphics[width=0.75\columnwidth]{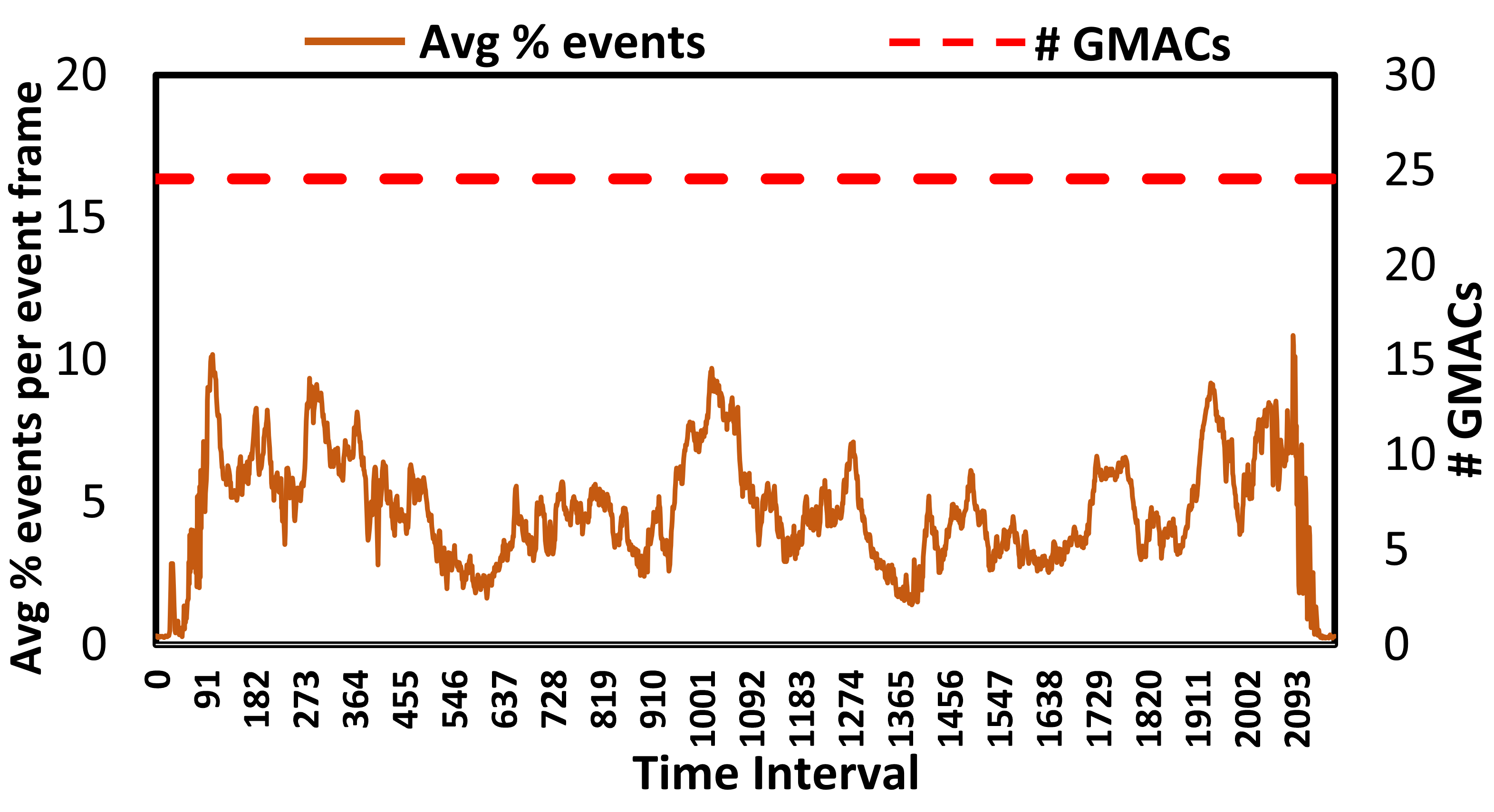}
  \vspace*{-20pt}
  \caption{Average percentage of events in an event frame and the number of operations expanded for processing events: Adaptive-SpikeNet~\cite{adaptive} on Multi Vehicle Stereo Event Camera (MVSEC) Indoorflying1 dataset~\cite{mvsec}}
  \label{fig:motiv}
  \vspace*{-6pt}
\end{floatingfigure}

Commodity edge platforms such as NVIDIA Jetson~\cite{jetson} and Google Coral~\cite{coral}, equipped with heterogeneous processing elements like CPUs, GPUs and specialized neural network accelerators (DLA/TPU/NPU) have gained prominence in running machine learning workloads efficiently. However, we observe that deploying event-based vision systems on these heterogeneous edge platforms leads to suboptimal performance. This is due to the fundamental differences between the sensing modality and the processing requirements of the network on the one hand, and the underlying hardware platform on the other. For example, these networks often convert the raw event streams to image-like event frames. Subsequently, these event frames are processed with fixed-sized matrix multiplication operations regardless of the the number of events generated, resulting in a significant proportion of redundant and wasteful operations, as shown in Figure ~\ref{fig:motiv}. Although sparse routines could be applied to the event frames, encoding and decoding overheads outweigh the potential benefits. Another crucial factor is the inability of the hardware platform to match the rate at which the vision sensor emits event data. Existing approaches ~\cite{fusionflow,spikeflow} either construct event frames by statically counting events or sampling events at a fixed rate without considering the hardware processing capabilities.  Finally, in many applications, there is a need to concurrently execute multiple networks in a manner that efficiently shares the resources in the hardware platform. 


Our work, ~\name{}, addresses these challenges and improves the performance of event-based algorithms on heterogeneous edge platforms by introducing three key optimizations. First, we propose an Event2Sparse Frame converter (E2SF) to convert the raw event streams directly to a sparse frame representation eliminating the need for intermediate event frames. This optimization enables the use of sparse libraries and ensures that the computational overhead is proportional to the number of generated events. Next, we present Dynamic Sparse Frame Aggregator (DSFA), an optimization that improves hardware utilization by dynamically merging sparse frames based on input dynamics and hardware processing capabilities. Finally, we propose Network Mapper (NMP) to allocate concurrently executing tasks to different processing elements while also optimizing layer precision considering both compute and communication overheads.

Across several state-of-the-art ANNs, SNNs and hybrid SNN-ANNs, ~\name{} achieves 1.28x-2.05x improvements in latency and 1.23x-2.15x in energy over an all-GPU implementation on the NVIDIA Jetson AGX Xavier for single task scenarios. Moreover, ~\name{} also achieves 1.43x-1.81x latency improvements over round-robin scheduling methods in multi-task execution scenarios with negligible accuracy loss.

\vspace*{-3pt}

\section{Background}
\noindent\textbf{Event cameras and Input representation.} Event cameras are neuromorphic sensors that operate at a high temporal resolution by capturing changes in brightness (log intensity) asynchronously for every pixel in a scene. When the log intensity ($I$) at a particular pixel is greater than the threshold ($\theta$) (i.e., $||log(I(t+1)) \minus{} log(I(t))|| >= \theta$), the camera generates an event. The event data is emitted in Address Event Representation (AER) format as \{x,y,t,p\}, where x and y corresponds to the pixel location, t represents the timestamp and p is a binary value that indicates whether the polarity of the brightness change is positive or negative.

\begin{figure}[h]
  \centering
  \vspace*{-12pt}
  \includegraphics[width=\linewidth]{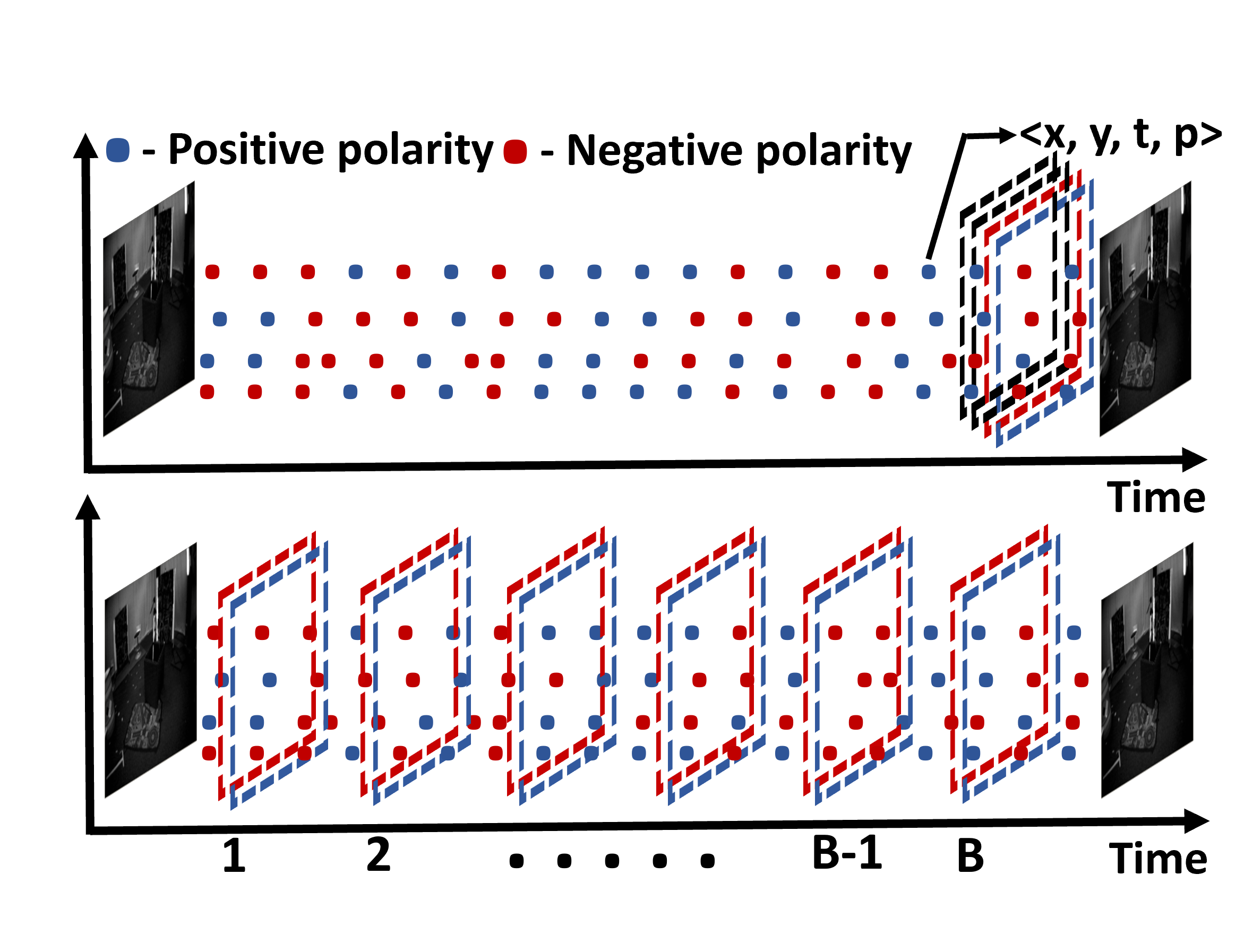}
  \vspace*{-25pt}
  \caption{Popular input representation schemes employed by state-of-the-art event-based algorithms}
  \label{fig:inprep}
  \vspace*{-12pt}
\end{figure}
For processing efficiency and network compatibility, as shown in Figure~\ref{fig:inprep}, the asynchronous events are accumulated over a time period to construct event frames. Some works~\cite{evflow} fully accumulate events between two consecutive image frames and encode the total number of events along with the most recent timestamp of intensity change in a pixel. Other works such as~\cite{depth,spikeflow} discretize events between two consecutive image frames into uniformly separated synchronous event frames. During inference, these event frames are either presented to the network as a single input with B channels or sequentially over B/k timesteps, where k denotes the number of concatenated event frames in a timestep. We note that ~\name{} supports all of the aforementioned input representations.

\vspace*{-3pt}

\section{Related Work}
In this section, we review prior related research efforts and compare ~\name{} with them.

\noindent\textbf{Algorithmic Techniques.} In order to exploit the spatio-temporal sparsity of events, ~\cite{sparseconv1} and ~\cite{sparseconv} introduce novel methods for computing event-based convolutions. Our work complements these efforts to provide additional performance gains.

\noindent\textbf{Hardware accelerators.} Specialized accelerators such as ~\cite{sne,kraken} accelerate key computations for processing asynchronous event data. These efforts are orthogonal to our work since we propose algorithmic optimizations to improve the performance of event-based networks on off-the-shelf hardware platforms. Nevertheless, some of our optimizations are also applicable to specialized accelerators.

\noindent\textbf{Mapping frameworks.} Several efforts ~\cite{comb, h2h, magma} have formulated mapping of multiple ANNs to a heterogeneous platform as an optimization problem. They consider the cost of compute and communication during the search to determine the final mapping configuration. The Event2Sparse Frame converter and Dynamic Sparse Frame Aggregator components of ~\name{} are complementary to these frameworks and specifically target the unique challenges of processing temporal event streams. The Network Mapper in ~\name{}, while similar in objective to these mapping frameworks, also includes a search for layer precision as part of the mapping process.


\section{\name{} Framework}

\begin{figure*}[htb]
  \includegraphics[width=\textwidth]{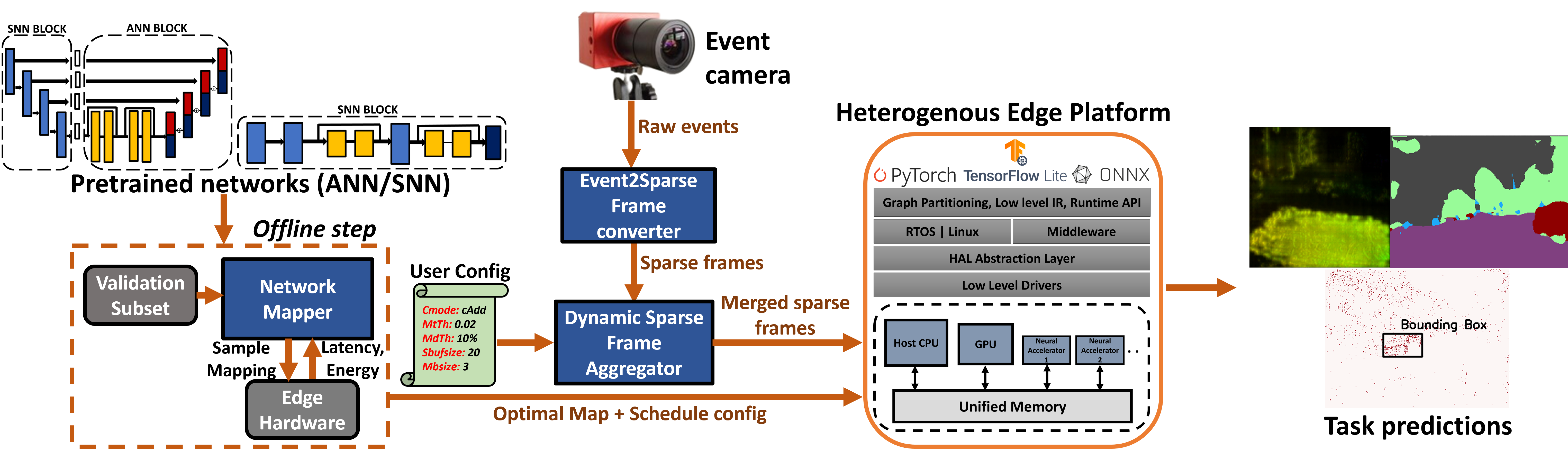}
  \caption{~\name{} framework for improving performance of event-based vision tasks on heterogeneous edge platforms. ~\name{} consists of three components,~\emph{viz.}, Network Mapper, Event2SparseFrame converter and Dynamic Sparse Frame Aggregator integrated into the inference pipeline}
  \label{fig:overall}
\end{figure*}

\name{} is an optimization framework designed to efficiently execute event-based vision algorithms on heterogeneous edge platforms. Figure~\ref{fig:overall} presents an overview of the ~\name{} framework. As an offline step, the pretrained networks are input to the Network Mapper (NMP) which determines the processing element to execute each layer on and the execution order. During inference, the event camera produces raw event streams that are directly converted to sparse frames by the Event2Sparse Frame converter (E2SF). Subsequently, the Dynamic Sparse Frame Aggregator (DSFA) merges these sparse frames while considering both the input dynamics and the hardware processing capabilities to generate merged sparse frames. It is important to note that NMP is an offline process that is executed only once, while E2SF and DEFA optimizations are applied at runtime during inference. The following subsections will delve into a comprehensive description of each of the components. 

\subsection{Event2Sparse Frame Converter (E2SF)}

\begin{floatingfigure}[r]{0.75\columnwidth}
  \centering
   \includegraphics[width=0.75\columnwidth]{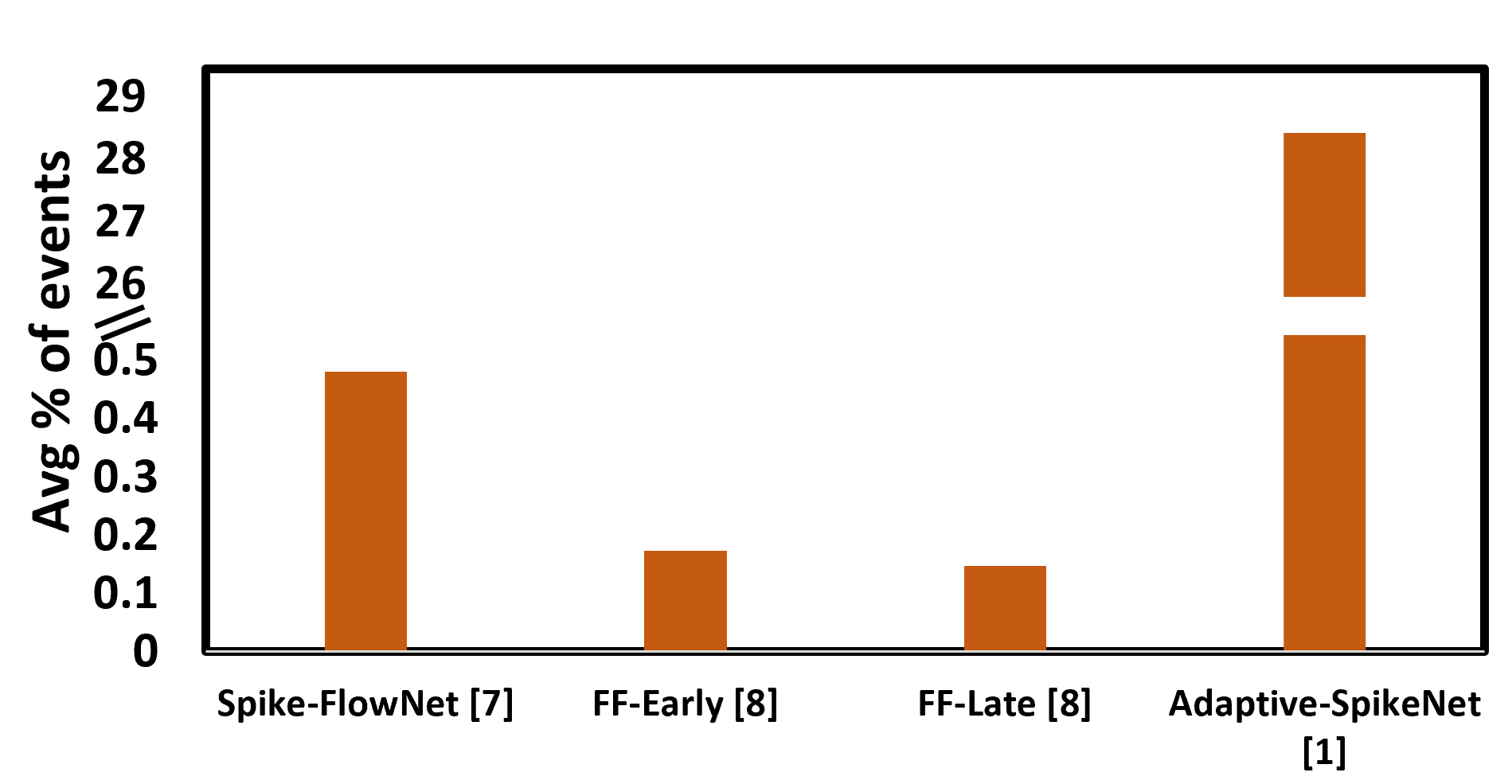}
  \vspace*{-20pt}
  \caption{Average percentage of events in each event frame for different networks on the MVSEC dataset}
  \label{fig:e2sf}
  \vspace*{-6pt}
\end{floatingfigure}

Several state-of-the-art event-based networks convert asynchronous raw event streams to a synchronous dense event frame representation of the data. As shown in Figure~\ref{fig:e2sf}, the average number of actual events per event frame can vary widely (0.15\%-28.57\%) across different networks that estimate optical flow~\cite{spikeflow, fusionflow,adaptive} on the MVSEC~\cite{mvsec} dataset, indicating that most event frames are extremely sparse. This results in a high number of wasteful computations as well as increased memory requirements. While the dense event frames can be converted to sparse tensors and processed with sparse linear algebra libraries, the associated encoding and decoding overheads are prohibitive. Instead, we propose an Event2Sparse Frame converter (E2SF) that converts the raw event stream from the camera directly to a sparse frame representation, enabling faster processing of the generated events on commodity edge platforms. 

Event cameras such as the Dynamic and Active Vision Sensor (DAVIS)~\cite{dvs} output synchronized grayscale frames along with the asynchronous event data. We refer to the timestamps of a pair of consecutive grayscale frames as \emph{Tstart} and \emph{Tend}. First, E2SF maps the event timestamps to discretized event bins based on Equation~\ref{eqn:bis}. The number of event bins (\emph{nB}) determines the temporal resolution of the data, which refers to how precisely information is captured in a scene. 
\vspace*{-6pt}
\begin{equation}
\begin{gathered}
\label{eqn:bis}
biS = (Tend - Tstart) / nB\\ 
EB_k = floor((t_k - Tstart)/biS) 
\end{gathered} 
\end{equation}
The duration of time interval assigned to each event bin, ~\emph{biS} is determined and used to scale each timestamp to its respective event bin index $EB_k$. Once the events are assigned to their respective event bins, we accumulate the positive and negative polarities separately for same pixel locations within each event bin. Finally, we iterate over the pixels in each accumulated event bin and store the row indices, column indices and their corresponding polarities as separate channels, similar to the sparse Coordinate (COO) format. Therefore, each event bin is converted to a two-channel sparse frame (positive and negative polarity), like ~\cite{spikeflow}. Converting the event data to sparse frame representations enables the use of sparse libraries ~\cite{submanifold} for improved performance during inference.

\subsection{Dynamic Sparse Frame Aggregator (DSFA)}

\begin{figure}[h]
  \centering
  \includegraphics[width=\linewidth]{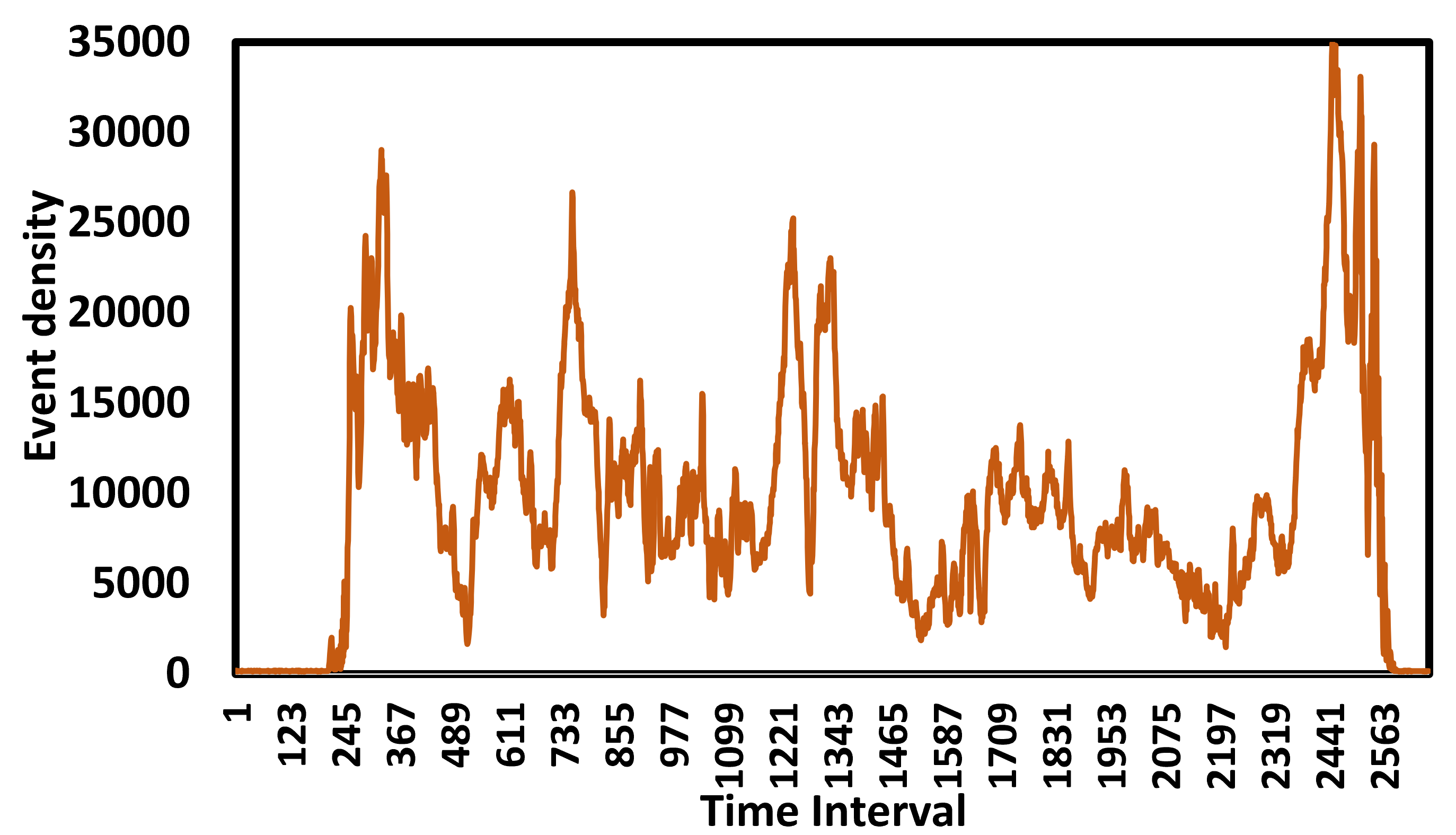}
  \vspace*{-23pt}
  \caption{Temporal event density of Indoorflying2 segment from MVSEC dataset~\cite{mvsec}}
  \label{fig:dsfamotiv}
\end{figure}

After E2SF transforms the raw events from the camera to synchronous sparse frames, the network processes these frames on the hardware platform to generate task predictions. However, we make a key observation that there still exists
a large variance in the number of events generated over time (temporal density) by the vision sensor (Figure ~\ref{fig:dsfamotiv}). Previous approaches~\cite{spikeflow,fusionflow} construct event frames statically either by counting the number of events that have occurred, or at fixed time intervals. However, they do not consider the hardware processing time while generating event frames, resulting in a backlog of event frames during periods of high activity that significantly increases overall latency. To address this challenge, DSFA merges the sparse frames at runtime by adapting to both the varying temporal event densities and the hardware processing capabilities, thereby improving latency with minimal accuracy loss. 


\begin{figure}[h]
  \centering
  \includegraphics[width=\linewidth]{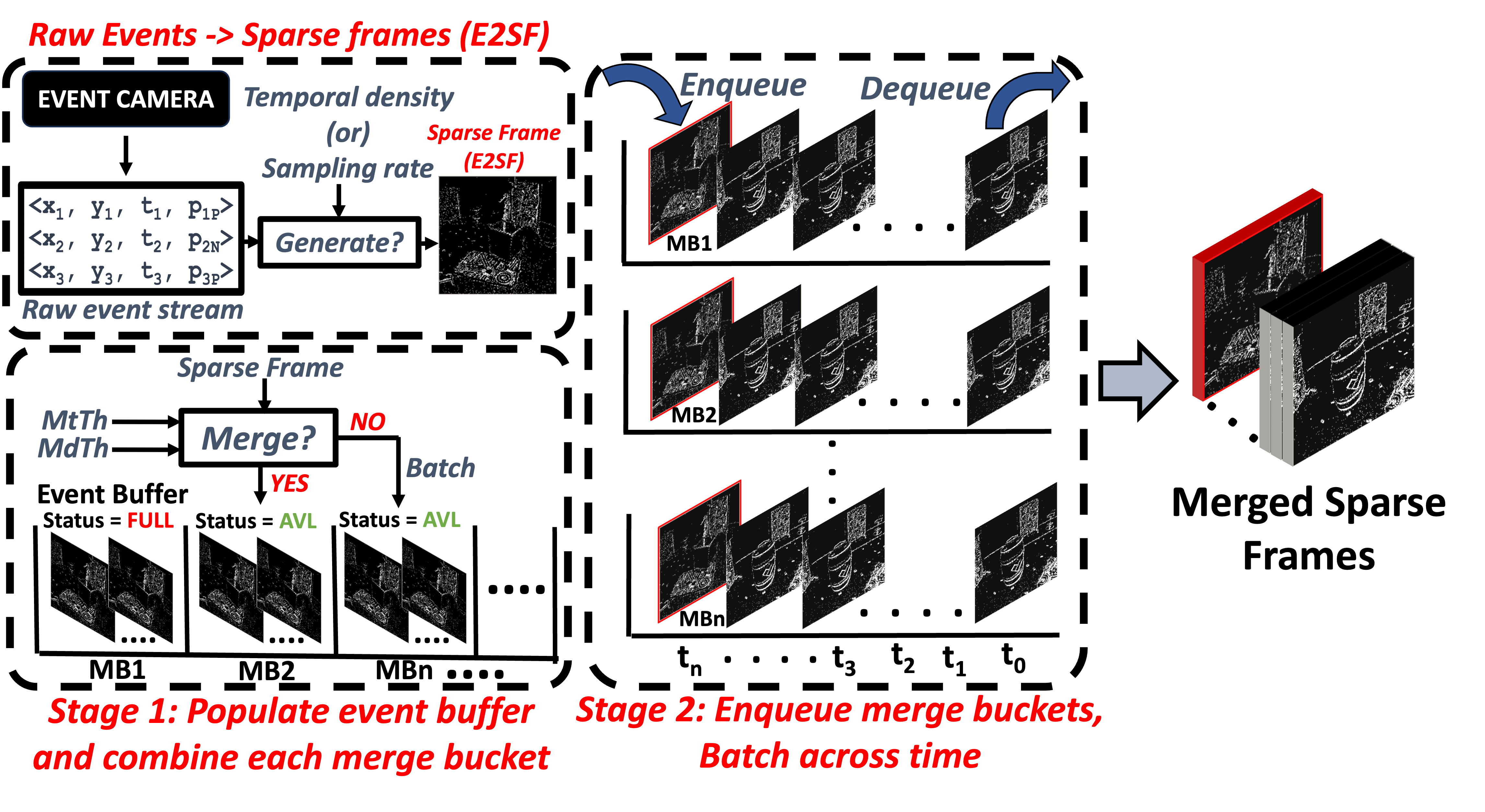}
  \vspace*{-23pt}
  \caption{Dynamic Sparse Frame Aggregator (DSFA)}
  \label{fig:dsfa}
\end{figure}

Figure ~\ref{fig:dsfa} outlines the DSFA procedure in detail. DFSA consists of two stages in order to generate merged sparse frames. We populate the event buffer first with sparse frames and propose a novel aggregation method to combine them. Subsequently, we dispatch the merged sparse frames to an inference queue which is later combined to form a batched input representation.

\noindent Initially, we allocate an event buffer of size \emph{EBufsize} to store the generated sparse frames(\emph{Evf\_k}). The event buffer is then partitioned into merge buckets of capacity \emph{MBsize}, each designated to contain a specific subset of sparse frames. These sparse frames are merged depending on the CMode parameter. Based on the task, DSFA supports three merging modes (\emph{cMode}); \emph{cAdd}, which adds the pixels across frames, \emph{cAverage}, that averages the pixels across frames, and \emph{cBatch}, which concatenates the sparse frames. For instance, in high-speed scenarios, it is recommended to use \emph{cBatch} mode to capture the scene accurately. Every merge bucket (MB) tracks the current capacity, time of earliest spike frame (\emph{Time(Evf\_1)}) and merged bucket density (\emph{MBmerged}). Furthermore, each merge bucket has a status flag that indicates whether it can accept new event frames (\emph{AVL}) or is at maximum occupancy (\emph{FULL}). When a sparse frame \emph{Evf\_k} is generated, we use a greedy approach to place it in the earliest available bucket, subject to two conditions (i) The time delay between the current \emph{Evf\_k} and the earliest entry in the merge bucket is within the maximum delay threshold, \emph{MtTh} (ii) The percentage change in the spatial density of the current event frame and \emph{MBmerged} is less than the maximum density threshold, \emph{MdTh}. We note that both \emph{MtTh} and \emph{MdTh} needs to be tuned for each task individually. If \emph{Evf\_k} fails to meet either of these conditions, the status of the current bucket is changed to \emph{FULL}, and the sparse frame is placed in the next merge bucket. However, when \emph{cMode} is set to \emph{cBatch}, every generated \emph{Evf\_k} is placed in a new merge bucket. If the event buffer occupancy is greater than \emph{EBufsize}, we combine each merge bucket according to the specified \emph{CMode}. The resulting merge buckets are forwarded to their respective inference queues as the latest sparse frames, where the earliest sparse frames in each queue is discarded. The merge buckets are concatenated to form a merged sparse frame representation. This helps in improving the hardware utilization and efficiency of inference execution due to batching of the sparse frames. If the hardware platform becomes available before the event buffer reaches full capacity, we dispatch the available merge buckets to maximize hardware utilization. It is also important to choose an optimal \emph{MBsize} to achieve the best tradeoff between accuracy and performance. For instance, a smaller \emph{MBsize} captures temporal information better thereby representing the scene more accurately while a larger \emph{MBsize} consolidates the events to fewer sparse frames leading to improved performance.
\vspace*{-10pt}
\subsection{Network Mapper (NMP)}
\vspace*{-2pt}
\begin{figure}[h]
  \centering
  \includegraphics[width=\linewidth]{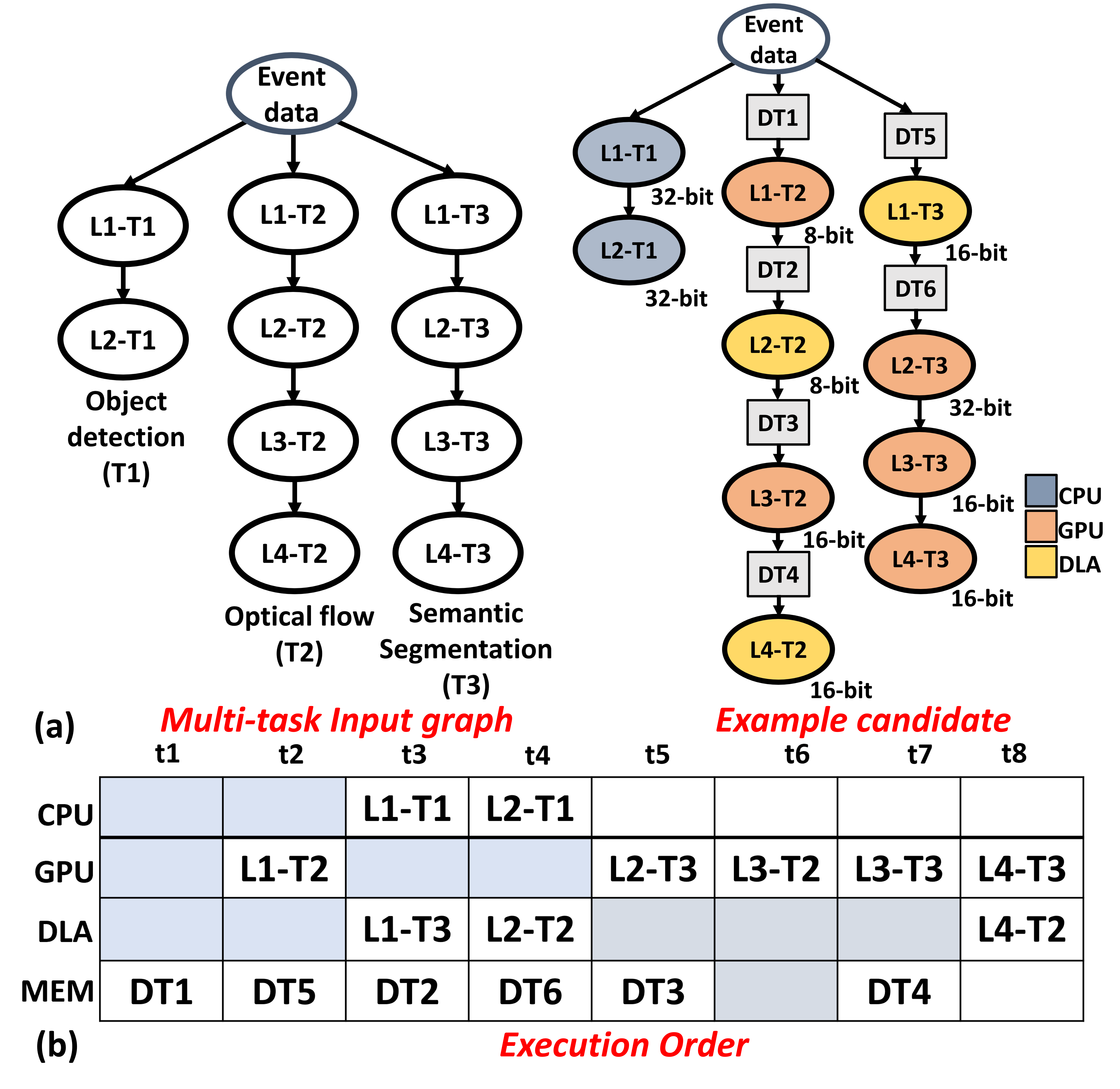}
  \vspace*{-16pt}
  \caption{(a) Example multi-task input graph. Candidate generated from the input graph by assigning each node to a processing element and adding communication nodes between nodes, (b) Execution order of the example candidate}
  \label{fig:nmp}
  \vspace*{-12pt}
\end{figure}
\noindent NMP efficiently maps the layers of multiple networks to different processing elements on heterogeneous edge platforms while also selecting the optimal layer precision during concurrent execution. The optimization process takes into account the computational and communication overheads and the accuracy degradation due to mixed precision. We represent the multi-task network dependencies as a directed graph, illustrated in Figure~\ref{fig:nmp}(a). Each node in the multi-task graph corresponds to a distinct layer of a network, while each edge signifies the data dependencies between the corresponding layers. Within the multi-task input graph, each node can be mapped to a variety of processing elements \{$D_1$, $D_2$, ..., $D_k$\} of the heterogeneous platform. Additionally, each processing element offers a range of precision options. The possible configurations encompassing all feasible mapping schemes and precision choices scales exponentially with the number of layers, i.e., $(\#Precision Choices * \#Processing Elements) ^ {\#Layers}$.  Hence, relying on naïve ad-hoc methods is impractical. Instead, we formulate the design space exploration as an evolutionary search problem to determine the best mapping and layer precision. We define the objective function in Equation~\ref{eqn:nmpformul} to minimize the latency of tasks \{$T_1$, $T_2$, ..., $T_n$\} such that the accuracy degradation for each task \{$\Delta A_1$, $\Delta A_2$, ...,  $\Delta A_n$\} due to mixed precision is within a defined threshold $\Delta$ A. Similarly, this procedure can be repeated to optimize for other objectives such as energy as well. We now describe our mapping methodology and latency estimation in more detail in sections ~\ref{sec:mapmethod} and ~\ref{sec:schedmethod} . 

\begin{equation}
\begin{gathered}
\label{eqn:nmpformul}
\min(\max_{i=1..n}Latency(T_i))\hspace{2pt} s.t. \hspace{2pt}\Delta A_1,\Delta A_2...\Delta A_n<=\Delta A \\
\Delta A_n = || Accuracy_{base} - Accuracy_{search} || \\
Mapping\hspace{2pt}  \rightarrow  \hspace{2pt}(D_1, D_2 ... D_k)
\end{gathered} 
\end{equation}

\subsubsection{Mapping methodology}
\label{sec:mapmethod}
We convert the multi-task input graph to a potential candidate (mapping configuration) by assigning each layer to a processing element and a precision choice supported by the processing element, as shown in Figure~\ref{fig:nmp}(a). We also insert data transfer nodes between layers to account for the communication overhead whenever the producer-consumer layers of a task are assigned to different processing elements. We first choose a random initial candidate set of potential solutions. After the initial population is chosen, each candidate is evaluated based on a fitness score. Subsequently, fitter generations of parents are created by crossover and mutation. This procedure is repeated for several iterations to choose the best mapping and precision for the network. Below are the key steps of our evolutionary search algorithm.\\ 
\noindent\textbf{Candidate evaluation.} The fitness of each candidate is evaluated using the objective function presented in Equation~\ref{eqn:nmpformul}. To compute the accuracy of each task, the pretrained network is quantized linearly based on the layer bit-widths specified in the candidate set and evaluated on a validation dataset. It is important to note that accuracy is determined individually for each task. The latency is estimated based on the execution order of the layers on the hardware platform, which is determined by an efficient scheduling algorithm described in section ~\ref{sec:schedmethod}. This process is repeated for all candidates, and their respective fitness scores are stored. Since the search spans multiple iterations, we employ certain  optimizations techniques to reduce its complexity. First, we perform inference only on a randomly sampled subset of the validation set. Second, the fitness scores are cached for each new candidate and reused if the same candidate emerges from different parents.

\noindent\textbf{Crossover.} After the parents are chosen, new children are produced by the fittest candidates in the current generation. For every pair of neighboring parents, one of them is chosen as the child for the next generation randomly based on equal likelihood. 

\noindent\textbf{Mutation.} A specified number of layers in each task is replaced with a random mapping resource and precision choice.
 
\subsubsection{Scheduling methodology and Latency Estimation}
\label{sec:schedmethod}
\noindent Calculating the fitness score for each candidate requires determining the
execution order of the layers and estimating latency. This procedure needs to be fast since it needs to be repeated for a large number of candidates across multiple generations.To that end, as shown in Figure ~\ref{fig:nmp}(b), we establish an execution queue for each device including unified memory in the heterogeneous platform to facilitate parallel execution of certain layers. Based on the data dependencies in the multi-task input graph, we obtain a partial ordering of layers. Subsequently, using this ordering, we serialize nodes within their respective execution queues (obtained from the mapper) that are not already serialized by the data dependencies. As shown in Equation~\ref{eqn:schedul}, the end time for each node, \emph{End\_T(node)} is determined based on three factors,~\emph{viz.}, the completion of all its parent nodes ~\emph{(End\_T(ParentN))}, the actual execution time of the node ~\emph{(Exec\_T(Node))} and completion of all the nodes preceding it in the execution queue ~\emph{(CurDeviceQ\_T)}. Once the end times for all nodes in the graph have been determined, the latency of the candidate is calculated by analyzing the critical path of the graph. The individual execution time for each layer and the communication time between layers are measured on the hardware platform and recorded before the search process begins.

\small

\begin{equation}
\begin{gathered}
\label{eqn:schedul}
End\_T(Node)=max(End\_T(Parent1)...End\_T(ParentN)\\
                  \hspace{3.2cm},CurDeviceQ\_T)+Exec\_T(Node)\\
CriticalPathLatency=max(End\_T(Node1)...End\_T(NodeN))
\end{gathered} 
\end{equation}

\normalsize

\vspace*{-6pt}

\section{Experimental Methodology}
In this section, we discuss the experimental setup and benchmarks for evaluating ~\name{}.\\
\noindent\textbf{Applications and Datasets.} We evaluate ~\name{} across different tasks and networks summarized in Table ~\ref{tab:networkeval}. We note that optical flow, semantic segmentation and object tracking tasks are validated on different sequences of the Multi Vehicle Stereo Event Camera (MVSEC) dataset ~\cite{mvsec} while the depth estimation task is validated on the Town 10 sequence of Depth Estimation oN Synthetic Events (DENSE) dataset ~\cite{depth}. During multi-task execution, we evaluated our mapping methodology across an all-ANN (~\cite{evflow},~\cite{depth}), all-SNN (~\cite{dotie}, ~\cite{adaptive}) and mixed SNN-ANN configurations (~\cite{fusionflow}, ~\cite{halsie}, ~\cite{dotie}, ~\cite{depth}).\\
\noindent\textbf{Simulation Setup.} ~\name{} is developed using PyTorch and evaluated on the Jetson Xavier AGX board, a heterogeneous edge platform equipped with a CPU, GPU, and DLA for accelerating deep learning workloads. To profile layer execution times at various precision levels, we use TensorRT ~\cite{tensorrt} on both the GPU and DLA. Since there is no explicit method to measure communication times between layers, we approximate the overheads based on the amount of bandwidth and data volume. We utilize the \textit{Tegrastats} tool to obtain power measurements.

\small
\begin{table}
\caption{Summary of Networks}
\centering

\begin{center}
\begin{tabular}
{|P{20mm}|P{20mm}|P{15mm}|P{15mm}|}
\hline
\textbf{Network} &  \textbf{Task} & \textbf{Type} & \textbf{\# Layers} \\
\hline
SpikeFlowNet~\cite{spikeflow} &Optical Flow &SNN-ANN &\textbf{12} (4 SNN, 8 ANN) \\
\hline
Fusion-FlowNet~\cite{fusionflow} &Optical Flow &SNN-ANN &\textbf{29} (10 SNN, 19 ANN) \\
\hline
Adaptive-SpikeNet~\cite{adaptive} &Optical Flow &SNN &\textbf{8} \\
\hline
HALSIE~\cite{halsie} &Semantic Segmentation& SNN-ANN &\textbf{16} (3 SNN, 13 ANN) \\
\hline
J. Hidalgo-Carrio et. al~\cite{depth} &Depth Estimation &ANN &\textbf{15}  \\
\hline
DOTIE~\cite{dotie} &Object Tracking &SNN &\textbf{1} \\
 
\hline

\end{tabular}
\end{center}
\label{tab:networkeval}
\end{table}

\normalsize

\vspace*{-6pt}

\section{Results}
This section illustrates the performance improvements obtained by ~\name{} on state-of-the-art event-based networks.

\begin{figure}[h]
  \centering
  \includegraphics[width=\linewidth]{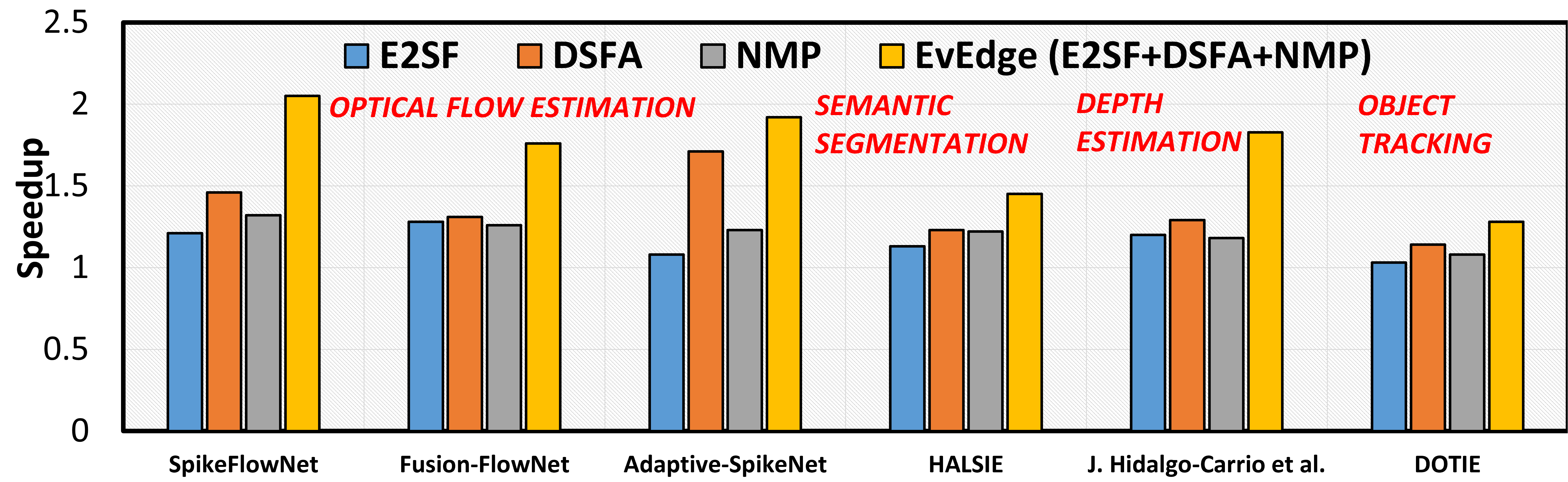}
  \vspace*{-16pt}
  \caption{Speedup results compared to an all-GPU implementation for single task execution.}
  \vspace*{-12pt}
  \label{fig:singletask}
\end{figure}

\noindent\textbf{Single-task execution performance.} Figure ~\ref{fig:singletask} shows the speedup results of ~\name{} compared to an all-GPU implementation on the Jetson Xavier AGX hardware platform. We present the improvements achieved by ~\name{} after each optimization is applied individually as well as the combined improvements. We observe that ~\name{} consistently outperforms the GPU implementation by 1.23x-2.05x across different levels of optimization. Notably, SNNs (depending on the number of layers) achieve the highest performance improvements, as they have the longest execution times on these platforms as well as high activation sparsity. Also, DSFA does not produce significant benefits in semantic segmentation networks due to the pixel-wise accuracy requirements, limiting the aggressiveness of merging frames. We also report the minimal accuracy degradation due to ~\name{} in Table ~\ref{tab:accdeg}. In addition to speedup improvements, ~\name{} also achieves 1.23x-2.15x energy efficiency improvements over an all-GPU implementation. 

\begin{figure}[h]
  \centering
  \includegraphics[width=\linewidth]{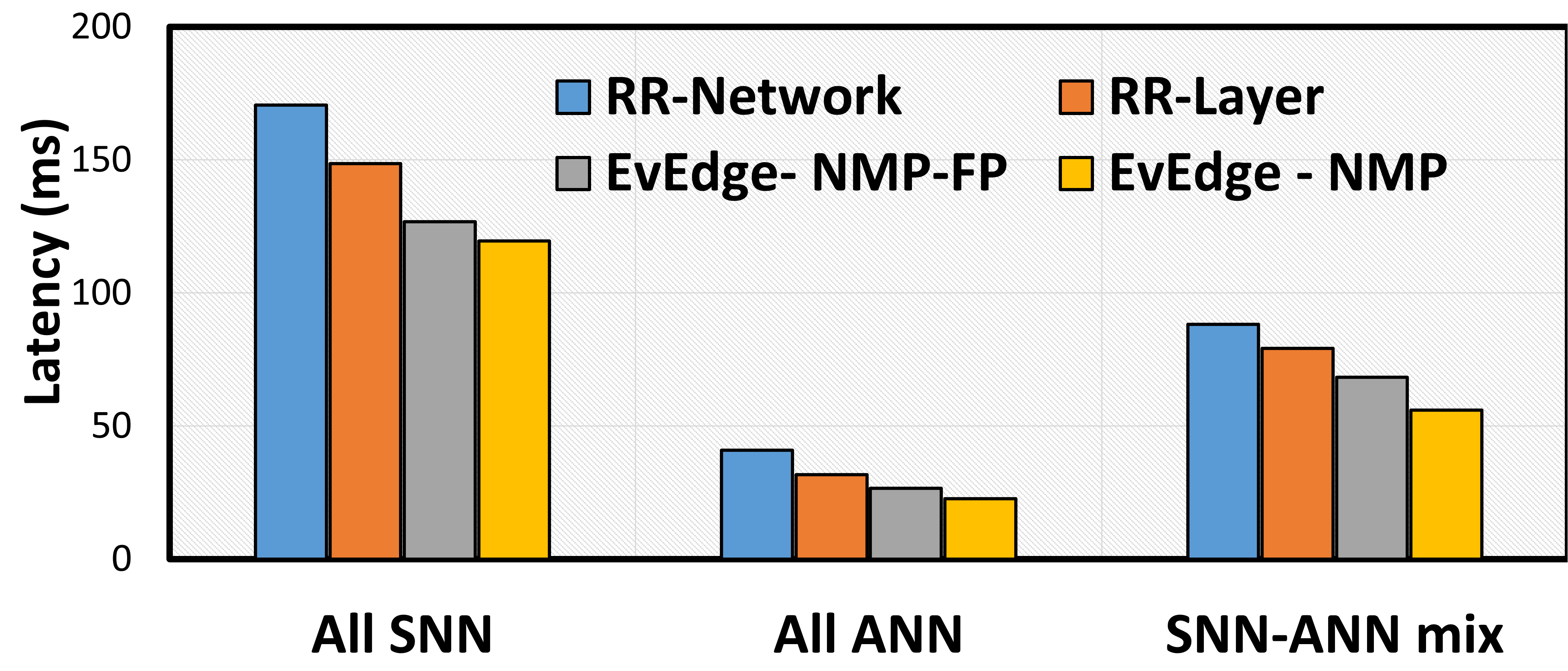}
  \vspace*{-16pt}
  \caption{Speedup results for multi task execution.}
  \vspace*{-12pt}
  \label{fig:multitask}
\end{figure}

\noindent\textbf{Multi-task execution performance.} Figure ~\ref{fig:multitask} shows the latency improvements of NMP for different multi-task network configurations. We compare ~\name{} to two variations of round-robin scheduling methods. RR-Network is a coarse-grained round-robin allocation policy where each network is assigned to a processing element and the rest of the networks are distributed in a cyclic manner. RR-Layer is a fine-grained round-robin allocation policy where each layer is assigned to a processing element. We notice that NMP provides 1.24x-1.41x improvements over RR-Layer and 1.43x-1.81x over RR-Network. This is because NMP searches from a much larger design space to optimize for both computation and communication overheads. We note that there is minimal accuracy degradation, since we are concurrently optimizing both accuracy and latency similar to single task execution. We also consider another variation of the network mapper, Ev-Edge-NMP-FP which exclusively maps to full precision cores to prevent any accuracy degradation. Ev-Edge-NMP-FP is 1.05x-1.22x slower than Ev-Edge-NMP but still significantly faster than the RR-Network and RR-Layer baselines. Ev-Edge-NMP-FP can be useful for accuracy sensitive applications.

\begin{figure}[h]
  \centering
  \includegraphics[width=\linewidth]{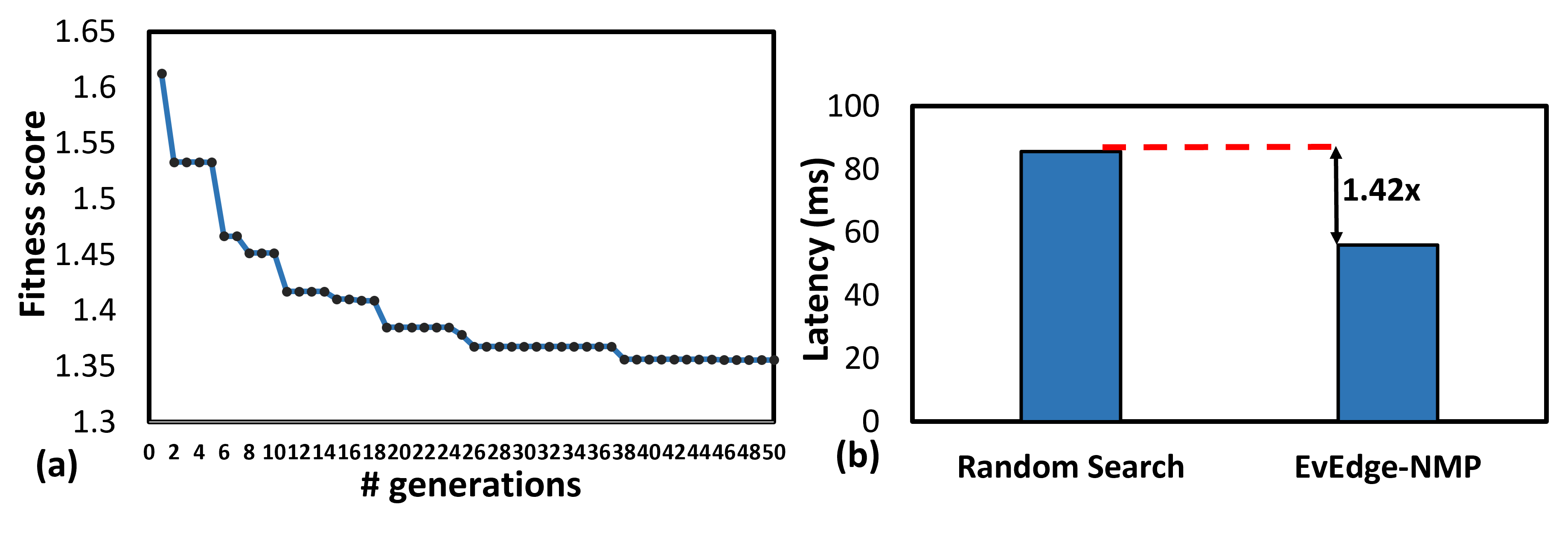}
  \vspace*{-21pt}
  \caption{(a) NMP evolutionary search convergence., (b) Latency of NMP searched configuration compared to random search for SNN-ANN mix configuration}
  \vspace*{-12pt}
  \label{fig:nmpsupp}
\end{figure}

Next we show the fitness score convergence over several generations of evolutionary search in Figure ~\ref{fig:nmpsupp}(a), indicating that latency and accuracy degradation are minimized simultaneously. We also compare NMP's evolutionary search technique in Figure~\ref{fig:nmpsupp}(b) to random search where the candidates are randomly sampled every generation, for a mixed SNN-ANN network. We observe that Ev-Edge-NMP is 1.42x faster compared to the results of random search.

\small
\begin{table}
\caption{Accuracy for single Task execution (Arrow indicates whether lower or higher is better)}
\centering

\begin{center}
\begin{tabular}{ |c|c|c| } 
\hline
\textbf{Network-Metric} &  \textbf{Baseline} & \textbf{Ev-Edge}\\
\hline
SpikeFlowNet~\cite{spikeflow} (AEE-$\downarrow$) &0.93 &0.96 \\
\hline
Fusion-FlowNet~\cite{fusionflow} (AEE-$\downarrow$)&0.72 &0.79 \\
\hline
Adaptive-SpikeNet~\cite{adaptive} (AEE-$\downarrow$) &1.27 &1.36 \\
\hline
HALSIE~\cite{halsie} (mIOU-$\uparrow$) &66.31 & 64.18  \\
\hline
J. Hidalgo-Carrio et al.~\cite{depth} (Avg Error-$\downarrow$) &0.61 &0.63  \\
\hline
DOTIE~\cite{depth} (mIOU-$\uparrow$) &0.86 &0.82\\
 
\hline

\end{tabular}
\end{center}
\label{tab:accdeg}
\end{table}

\normalsize


\vspace*{-6pt}
\vspace*{-0pt}

\section{Conclusion}
In this work, we introduced ~\name{}, a framework to improve the performance of event-based algorithms on heterogeneous edge platforms. ~\name{} consists of three optimizations that are integrated into the inference pipeline. First, E2SF directly transforms raw event streams to sparse frames without the need to convert to intermediate event frames. Next, DSFA combines the sparse frames at runtime by considering the properties of the event streams as well the hardware efficiency. Finally, NMP determines the best mapping configuration and layer precision for concurrently executing tasks. We demonstrated the efficacy of ~\name{} across several state of the art SNNs, ANNs and hybrid SNN-ANNs for both single task as well as multi-task scenarios.

\vspace*{0.1in}\noindent\textbf{Acknowledgements.} This work was supported in part by the Micro4AI program from IARPA and the Center for the Co-Design of Cognitive Systems (COCOSYS), a DARPA-sponsored JUMP center, Semiconductor Research Corporation (SRC).

\vspace*{-3pt}


\bibliographystyle{ACM-Reference-Format}
\bibliography{sample-base}

\end{document}